\pdfoutput=1
\PassOptionsToPackage{table, dvipsnames}{xcolor}
\documentclass[sigconf]{acmart}
\usepackage{booktabs, multirow}
\usepackage{amsmath}
\usepackage{xcolor}
\usepackage{xspace}
\usepackage{tcolorbox}
\usepackage[table]{xcolor}
\usepackage{subfigure}
\usepackage{fix-cm}
\usepackage{enumitem}
\usepackage{setspace}
\usepackage{tikz}
\usetikzlibrary{tikzmark}
\usepackage{float}
\usepackage{dsfont}
\usepackage{pifont}
\usepackage{bbding}
\usepackage{graphicx}
\usepackage{subfigure}
\usepackage{subcaption}

\definecolor{yellow}{rgb}{0.74, 0.56, 0}
\definecolor{purple}{rgb}{0.32, 0.09, 0.98} %{0.01, 0.01, 0.51}
\definecolor{red}{rgb}{0.81, 0.09, 0.13}

\definecolor{blue}{cmyk}{0.95,0.0,0.2,0.2}
\definecolor{lightyellow}{cmyk}{0.01,0.0,0.2,0.01}
\definecolor{lightblue}{cmyk}{0.1,0.0,0.02,0.02}
\definecolor{viridis3}{RGB}{33, 145, 140} 
\definecolor{mypurple}{RGB}{236, 203, 250}

\tikzset{mycircled/.style={circle,draw=black,fill=mypurple,inner sep=0.1em,line width=0.1em, scale=0.8}} % Define a custom TikZ style

\newcommand{\hyperrag}{\texttt{Hypercube-RAG}\xspace}
\newcommand{\ul}{\underline}

\AtBeginDocument{%
  }

\setcopyright{acmlicensed}
\copyrightyear{2025}
\acmYear{2025}
\acmDOI{XXXXXXX.XXXXXXX}
\acmConference[Conference acronym 'XX]{Make sure to enter the correct
  conference title from your rights confirmation email}{June 03--05,
  2018}{Woodstock, NY}
\begin{document}

\title{Hypercube-Based Retrieval-Augmented Generation for Scientific Question-Answering}

\author{Jimeng Shi$^{1}$, Sizhe Zhou$^2$, Bowen Jin$^2$, Wei Hu$^2$, Runchu Tian$^2$, \\ Shaowen Wang$^2$, Giri Narasimhan$^1$, Jiawei Han$^2$}
\affiliation{%
\institution{$^1$Florida International University,  $^2$University of Illinois Urbana-Champaign}
  \country{}
}
\email{
{jshi008, giri}@fiu.edu,
{bowenj4, weih9, runchut2, shaowen, hanj}@illinois.edu
}

\renewcommand{\shortauthors}{Jimeng Shi et al.}

\begin{abstract}
Large language models (LLMs) often need to incorporate external knowledge to solve theme-specific problems.
Retrieval-augmented generation (RAG) has shown its high promise, empowering LLMs to generate more qualified responses with retrieved external data and knowledge.
However, most RAG methods retrieve relevant documents based on either sparse or dense retrieval methods or their combinations, which overlooks the essential, multi-dimensional, and structured semantic information present in documents.
This structured information plays a critical role in finding concise yet highly relevant information for domain knowledge-intensive tasks, such as scientific question-answering (QA).
In this work, we introduce a \textit{multi-dimensional (cube) structure}, \textbf{Hypercube}, which can index and allocate documents in a pre-defined multi-dimensional space.
Built on the hypercube, we further propose \textbf{Hypercube-RAG}, a novel RAG framework for precise and efficient retrieval.
Given a query, \hyperrag first decomposes it based on its entities, phrases, and topics along with pre-defined hypercube dimensions, and then retrieves relevant documents from cubes by aligning these decomposed components with corresponding dimensions. 
Experiments on three datasets across different domains demonstrate that our method improves response accuracy by $3.7\%$ and retrieval accuracy by $5.3\%$ over the strongest RAG baseline. It also boosts retrieval efficiency (speed) by one or two magnitudes faster than graph-based RAG.
Notably, our \hyperrag inherently offers explainability by revealing those underlying dimensions used for retrieval.
% In the end, \hyperrag achieves accuracy, efficiency, and explainability simultaneously for scientific QA tasks.
% The anonymous data and code are available at \textcolor{brown}{\emph{\url{https://anonymous.4open.science/r/HyperRAG/}}}.
The data and code are available at
\emph{\url{https://github.com/JimengShi/Hypercube-RAG}}.
\end{abstract}


% \ccsdesc[500]{Information systems~Data mining}

% Keywords. 
% A "teaser" image appears between the author and affiliation
% information and the body of the document, and typically spans the
% page.


\maketitle

\section{Introduction}
\label{sec:intro}
\begin{figure}[t!]
\centering
  \input{tables/case_study1}
  \vspace{-2.ex}
  \caption{Hypecube- vs.\ Semantic RAG: A case study} 
  \Description{Hypecube- vs.\ Semantic RAG: A case study}
  \label{fig:casestudy}
\vspace{-1mm}
\end{figure}
Large language models (LLMs) often suffer from hallucinations and factual inaccuracies, especially for scientific question-answering tasks \cite{augenstein2024factuality}.
To address this phenomenon, retrieval-augmented generation (RAG) has emerged as the de facto approach, incorporating external domain knowledge to generate contextually relevant responses \cite{huang2025survey}.
Despite recent advancements, they exhibit key limitations in theme-specific applications on which the fine-grained topics are highly focused.
For instance, retrieved documents are often semantically similar to the query but overlook specialized themes (e.g., terminology or nuanced contextual cues) prevalent in specific literature or reports.
% \cite{ding2024automated}.
Moreover, the retrieval process often suffers from inefficiency and limited transparency, which are critical in high-stakes domains, such as healthcare \cite{pujari2023explainable}, chemistry \cite{wellawatte2025chemlit}, environment \cite{wang2025remflow}, and law \cite{pipitone2024legalbench}.

% ============== sparse rag ==============
The conventional \textit{sparse lexical retrieval} ranks a set of documents based on the appearance of query terms \cite{kadhim2019term}. 
Due to its reliance on the exact token overlap, it precisely captures specific themes or topics, and the retrieval process is efficient and interpretable. However, it struggles to retrieve contextually related or paraphrased documents that require semantic understanding.
%
% ============== semantic rag ==============
To mitigate the above limitation, \textit{dense embedding retrievers} were proposed by computing the similarity between vector embeddings of query-document pairs \cite{karpukhin2020dense}. 
While contextual understanding is relatively improved, the hurdle of missing specific themes remains, causing semantically similar but off-topic retrievals \cite{krishna2021hurdles, kang2024improving}.
Figure \ref{fig:casestudy} shows an example where the Semantic-RAG returns documents related to (or the lack of) precipitation in Florida, but overlooks the critical location-specific information for ``Melbourne Beach''.
Moreover, retrieval in an embedding space makes it challenging to interpret why certain documents were selected \cite{ji2019visual}.
%
% ============== graph rag ==============
Additionally, a \textit{graph-based} RAG represents documents as graph structures, where nodes correspond to entities or concepts and edges capture their relationships \cite{peng2024graph}. 
They offer explanations of the retrieval process by outputting the traversed subgraph.
However, two limitations persist still.
First, traversing such subgraphs introduces a significant risk of information overload, as the expanding neighborhoods may include substantial amounts of irrelevant themes \cite{wang2024graph}.
Second, graph-based retrieval methods are prone to inefficiencies and considerable scalability bottlenecks \cite{devezas2021review, zhang2025survey, han2024retrieval}, primarily stemming from the vast number of nodes and edges in their underlying graph structures.

\begin{figure}[ht!]
\centering
  \includegraphics[width=1\columnwidth]{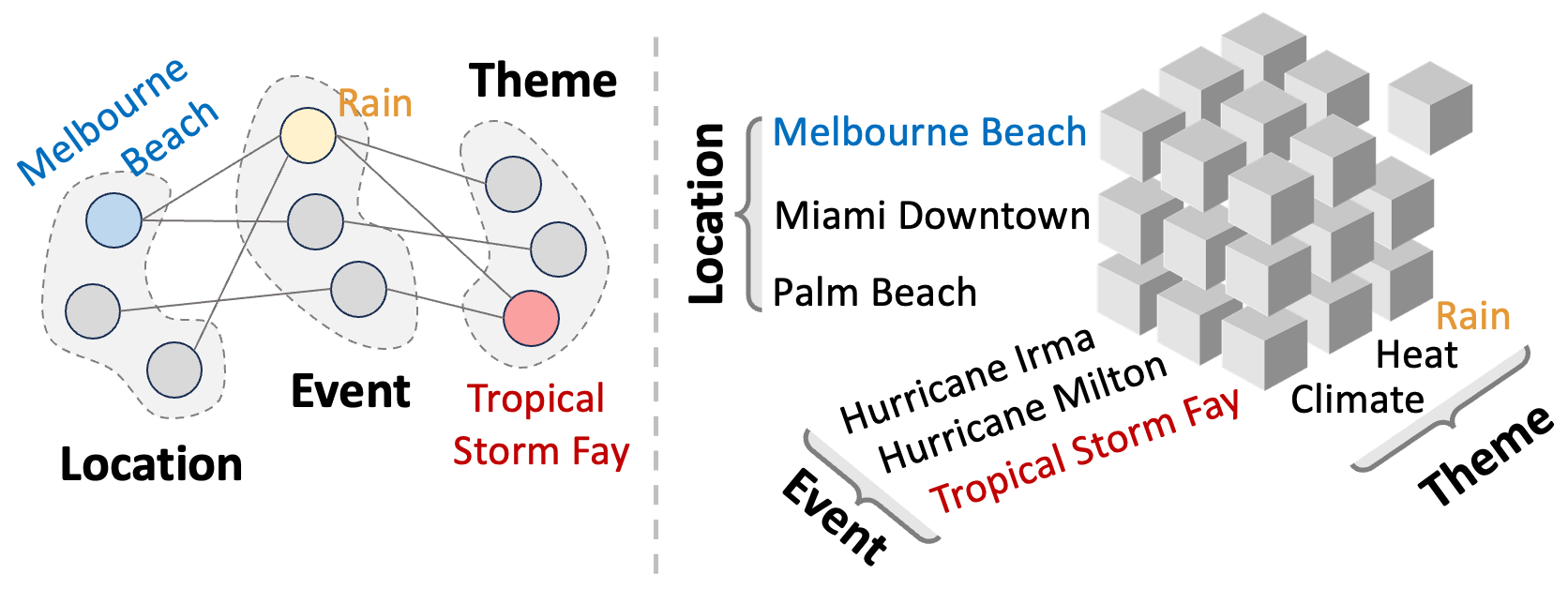}
  \caption{Structure comparison: Graph vs.\ Hypercube.} 
  \label{fig:graph_vs_hypercube}
\end{figure}

% ============== cube intro ==============
The methods discussed above tend to \textit{perform poorly in at least one of the aspects among accuracy, efficiency, and explainability.}
This motivates us to develop a RAG system that overcomes this weakness.
In this pursuit, we identified the \emph{text cube} as a promising technique \cite{tao2018doc2cube, wang2023geospatial}. 
A text cube is an inherently explainable multidimensional structure that allocates documents into cubes along various pre-defined dimensions, such as location, date, event, and specific-theme(s).
Each cell is associated with fine-grained entities (e.g., ``Miami Downtown'', ``2021'', ``Hurricane Irma'', ``flooding'') extracted from documents, which are viewed as document labels. 
Relevant documents can be indexed and located in cube cells associated with these document labels (cube labels for brevity).
Moreover, the theme dimension specifies the themes in hierarchies which make the text cube well-suited for \emph{theme-specific} applications where users primarily focus on \textit{fine-grained} inquiries \cite{tao2018doc2cube}, such as which hurricane hits and how much rainfall it results in certain areas, or which medicine is effective in treating which type of cancer.

\begin{figure}[H]
\centering
  \vspace{-2mm}
  \includegraphics[width=0.98\columnwidth]{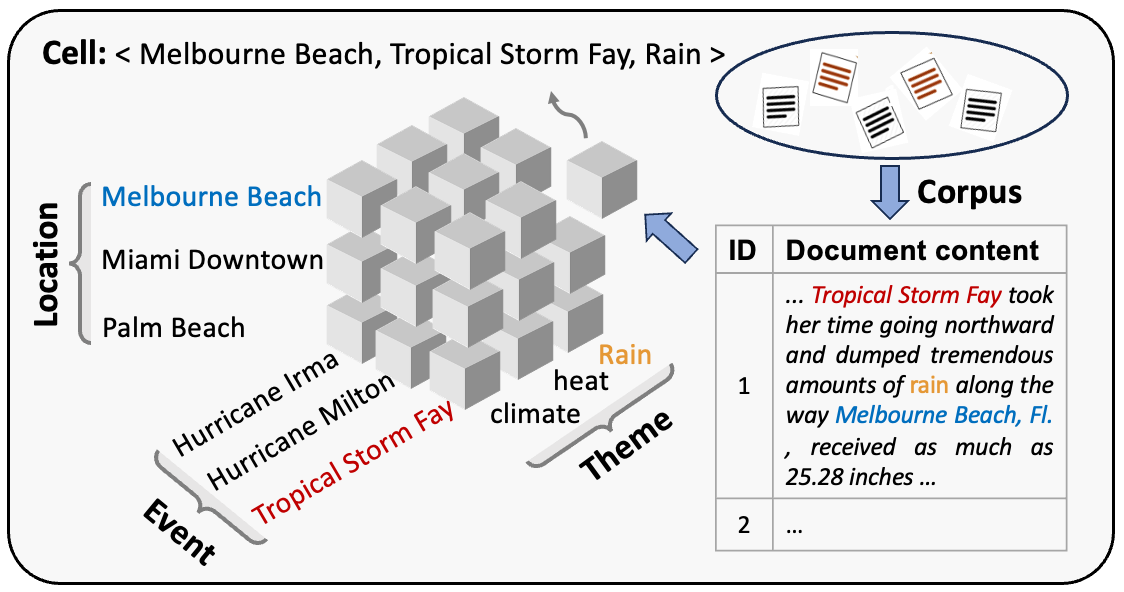}
  \caption{Hypercube construction on a corpus. We present only three dimensions for clear representation.} 
  \label{fig:hypercube}
\end{figure}

% ============== hypercube rag ==============
In this work, we implement a multi-dimensional (cube) structure called \textbf{Hypercube}, which indexes documents to cube cells based on their fine-grained labels in a pre-defined, multi-dimensional semantic space.
Building on this cube structure, we further propose \textbf{Hypercube-RAG}, a novel RAG framework to enhance the retrieval process by retrieving relevant documents from the corresponding cube cells.
Compared with the aforementioned RAG approaches,
our \hyperrag simultaneously exhibits three critical characteristics due to its multidimensionality:
(1) \textit{High Accuracy}. 
The multidimensional, cube label-based search supports both sparse and dense embedding strategies, effective for capturing essential information in multi-dimensional space as well as uncommon thematic terminology and semantically relevant cases.
(2) \textit{High Efficiency}.
The retrieval mechanism quickly narrows down the search to the pertinent cube cell(s) based on their associated labels. This operation exhibits a constant time complexity of $\mathcal{O}(c)$, which could be multiple magnitudes faster, thereby demonstrating potential robust scalability concerning corpus size.
% The retrieval process quickly narrows down the search to the right cube cell(s) based on the associated labels with the constant time $\mathcal{O}(c)$, which could be multiple magnitudes faster, demonstrating robust resilience to the corpus size.
% while taking only $\mathcal{O}(\kappa)$ to fetch $\kappa$ documents assigned in the corresponding cell(s).
(3) \textit{Explainability}. Hypercube is inherently explainable as fine-grained labels associated in cube cells represent the compact information in documents (see Figure \ref{fig:graph_vs_hypercube} (right) and Figure \ref{fig:hypercube}).
Overall, our contributions are as follows:
% \begin{itemize}[leftmargin=0.5cm,nosep]
\begin{itemize}[leftmargin=0.6cm]
    \item We identify the shortcomings of conventional RAG approaches, especially for \textit{theme-specific} question-answering tasks in the scientific domains.
    \item We design a multi-dimensional structure (hypercube) and further propose \hyperrag, a simple yet accurate, efficient, and explainable RAG method.
    \item We conduct experiments on three datasets across different domains with multiple metrics, demonstrating that our approach outperforms other baseline methods consistently.
\end{itemize}

\section{Related Work}
\label{sec:related}

\subsection{Text-based RAG}
BM25 \cite{robertson1994some, trotman2014improvements} is a classical sparse retrieval method based on TF-IDF principles, widely used in information retrieval. It ranks documents by scoring query term matches, with adjustments for term frequency saturation and document length. While efficient and interpretable, BM25 lacks semantic understanding and often misses paraphrased or contextually related documents without exact lexical overlap, limiting its effectiveness in tasks requiring deeper language comprehension.
Dense embedding retrievers compute contextual similarity between the given query and a document/chunk in the embedding space.
% \cite{jiang2023active}. 
Common retrievers consist of DPR \cite{karpukhin2020dense}, Contriever \cite{izacard2022contriever}, e5 \cite{wang2022text}, and ANCE \cite{xiong2020approximate}.
By capturing deeper semantic relationships, these methods improve performance in open-domain question answering and generative tasks, especially when surface-form matching fails.
% , such as with paraphrased queries or varied domain-specific terminology.
However, dense retrieval can return semantically close yet topically irrelevant documents, introducing noise and hallucinations. Moreover, its opaque nature also hinders interpretability \cite{ji2019visual}, limiting adoption in high-stakes or scientific domains. Recent efforts aim to mitigate these limitations by incorporating hybrid retrieval techniques \cite{maillard2021multi} and domain-adaptive retrievers \cite{lee2021learning}.

\subsection{Structured RAG}
Graph-based RAG methods enhance the standard RAG paradigm by introducing structured knowledge representations, such as entity or document graphs.
GraphRAG \cite{edge2024local} derives an entity knowledge graph from the source documents and gathers summaries for all groups of closely related entities. 
LightRAG \cite{guo2024lightrag} employs a dual-level retrieval system that enhances comprehensive information retrieval from both low-level and high-level knowledge discovery.
HippoRAG \cite{gutierrez2024hipporag} is a knowledge graph-based retrieval framework inspired by the hippocampal indexing theory of human long-term memory. 
HippoRAG 2 \cite{gutierrez2025rag} builds on top of HippoRAG and enhances it with deeper passage integration and more effective online use of an LLM. 
SELF-RAG \cite{asai2023self} enhances an LLM’s quality and factuality through retrieval
and self-reflection.
RA-DIT \cite{lin2023ra} fine-tunes the pre-trained LLMs or retrievers to incorporate more up-to-date and relevant knowledge.
Despite these improvements, graph construction and reasoning can be computationally intensive due to the vast number of nodes and edges.

\textit{Text Cube.}
The text cube has been proposed to structure spatial-related data across geospatial dimensions, thematic categories, and diverse application semantics \cite{wang2023geospatial}. 
STREAMCUBE implements the text cube to incorporate both spatial and temporal hierarchies \cite{feng2015streamcube}.
Doc2Cube \cite{tao2018doc2cube} automates the allocation of documents into a text cube to support multidimensional text analytics.
Despite these elaborate designs, the text cube has not yet been explored in the context of retrieval-augmented generation (RAG) systems.
\begin{figure*}[ht!]
\centering
  \includegraphics[width=1.99\columnwidth]{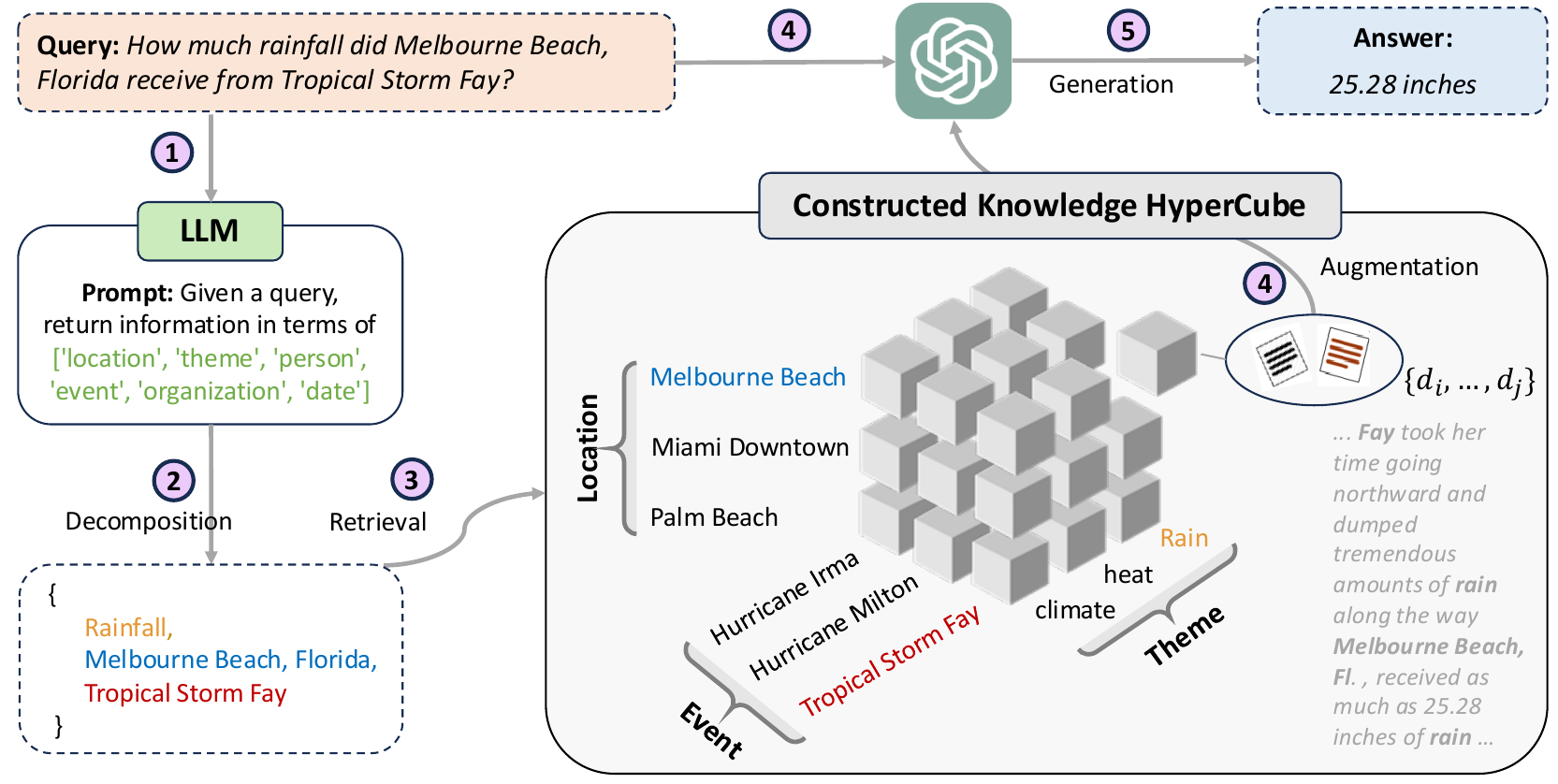}
  \vspace{-2mm}
  \caption{\hyperrag framework. 
      \tikzmarknode[mycircled]{a1}{1} \textbf{Input:} Input the query and the prompt into a LLM; 
      \tikzmarknode[mycircled]{a1}{2} \textbf{Decomposition:} LLM decomposes the query into different dimensions; 
      \tikzmarknode[mycircled]{a1}{3} \textbf{Retrieval:} according to these dimensions, we use \hyperrag to retrieve relevant documents; \tikzmarknode[mycircled]{a1}{4} \textbf{Augmentation:} query is augmented with retrieved documents (ranked already);
      \tikzmarknode[mycircled]{a1}{5} \textbf{Generation:} LLM output.
  } 
  \label{fig:hypercube_rag}
\end{figure*}

\section{Our Method: Hypercube-RAG}
\label{sec:hrag}
Figure \ref{fig:hypercube_rag} illustrates the \hyperrag framework through a concrete question-answering example.
In the following, we define the hypercube structure (Sec. \ref{sec:def_hyper}), describe the process for designing and constructing the hypercube (Sec. \ref{sec:construct_hyper}), explain how two sparse and dense embedding retrieval strategies are combined inside of hypercube (Sec. \ref{sec:retrieve_hyper}), and present how to rank retrieved documents (Sec. \ref{sec:rank_hyper}).

\subsection{Hypercube Formulation}
\label{sec:def_hyper}
Given a text corpus with documents $\mathcal{D}$, the hypercube is designed as a multidimensional structure, $\mathcal{C}_{ube}=\mathcal{C}_1\otimes \mathcal{C}_2 \otimes \dots \otimes \mathcal{C}_m$, where $m$ is the number of hypercube dimensions, and $\mathcal{C}_i$ is the $i^{th}$ dimension.
For any document $d \in \mathcal{D}$, a text classification algorithm (e.g, TeleClass \cite{zhang2025teleclass}) will assign it to one or more cube cells. 
This process is equivalent to assigning a $m$-dimensional labels ($c_1, c_2, \dots, c_m$) for a document, where label $c_j \in \mathcal{C}_j$ represents a category of document in $j^{th}$ dimension $\mathcal{C}_j$. 

\subsection{Hypercube Design and Construction}
\label{sec:construct_hyper}
\textbf{Hypercube Design.}
The core idea of the hypercube is to identify fine-grained entities or topics present in the corpus and to index documents into cube cells along appropriate dimensions, thereby enabling the subsequent retrieval. 
In this work, we design the hypercube structure by systematically determining its dimensions. 
First, we employ pre-trained language models to extract entities from the entire corpus.
Next, the K-means clustering algorithm is applied to group these entities into semantically coherent clusters.
% For example, countries, states, provinces, and specific places are clustered together within a location cluster, as we observed.
For each cluster, we let a large language model (LLM) generate concise summaries that characterize the underlying cluster categories or dimensions\footnote{For example, LLM outputted \textit{``these entities in the cluster include countries, states, provinces, cities, towns, and specific addresses.''}}.
Finally, we consolidate the candidate dimensions across clusters to construct a high-level, multi-dimensional space for document indexing.

This dimension-identification process for hypercube design has demonstrated strong alignment with domain expert expectations and yielded promising performance (although diverse strategies remain open for future exploration).
In Appendix \ref{sec:hypercube_design}, we provide a detailed dimension-identification process for each dataset. \\

\noindent \textbf{Hypercube Construction.}
With the hypercube dimensions finalized or pre-defined, we refine the previously generated concise summaries as prompts to an LLM to extract entities associated with each high-level dimension. 
These extracted entities are then used as labels to index documents into their corresponding cube cells. The process to extract entities for one dimension is represented as:
\begin{equation}
    % \mathcal{E}^{\texttt{EVENT}} = \{e\in \mathcal{E}(d) \mid \text{dim}(e)=\texttt{EVENT}\}.
    \mathcal{E}(d, \mathcal{C}_j) = \mathcal{F}(d, \text{prompt}(\mathcal{C}_j)),
\end{equation}
where $\mathcal{E}$ represents the extracted entity set from a document, $d$, along the $j^{th}$ dimension, $\mathcal{C}_j$, $\mathcal{F}$ refers to an LLM, and $\text{Prompt}(\mathcal{C}_j)$ denotes the corresponding prompt used for $\mathcal{C}_j$. We present a prompt example in Appendix \ref{sec:prompt}.

\subsection{Hypercube Retrieval}
\label{sec:retrieve_hyper}
A \textbf{constructed} knowledge hypercube can serve for the retrieval process based on its \textit{fine-grained labels}.
Given a query $q$, an LLM first decomposes it into a set of entities, $\mathcal{E}(q) = \{e_1, e_2, \dots, e_{l(q)}\}$, aligned with pre-defined dimensions (see the prompt in Figure \ref{fig:hypercube_rag}), then retrieval inside of hypercube is performed by matching these decomposed components with labels associated in cube cell(s).
Our hypercube supports two matching strategies: \textit{sparse exact lexical match} and \textit{dense embedding match}.
We prioritize leveraging the exact matching strategy due to its high precision, scoring documents based on the frequency with which query entities appear:
\begin{equation}
    \text{score}(d, q) = \sum_{e_j \in \mathcal{E}(q)} \mathds{1}_{[e_j \in \mathcal{E}(d)]},
\end{equation}
where $\mathds{1}$ is an indicator function that equals 1 if an entity of the query exactly matches a term in the document, and 0 otherwise.

Considering the example in Figure \ref{fig:hypercube_rag}, the last two components from the query, \textit{Melbourne Beach} and \textit{Tropical Storm Fay}, are handled using the exact matching strategy, as they correspond precisely to fine-grained labels in the hypercube.
However, the first component, \textit{Rainfall}, does not exactly match any label(s) in the hypercube, even though its matched label should be \textit{Rain}.

To address such cases, \hyperrag supports semantic retrieval by computing the similarity between query components and fine-grained cube labels, along with a certain dimension, $C_j$.
When the similarity score exceeds a predefined threshold $\tau$, semantic retrieval is triggered. 
Specifically, both the query entities and cube labels are projected into an embedding space, $\mathcal{Z}$, with an encoder. It can be represented as:
\begin{align}
    \mathcal{Z}(q) &= \text{Encoder}(\mathcal{E}(q)), \\
    \mathcal{Z}(d) &= \text{Encoder}(\mathcal{E}(d)), \\ 
    \text{score}(d, q) &= \text{sim}(\mathcal{Z}(q), \mathcal{Z}(d)),
\end{align}
where $\text{sim}(\cdot)$ refers to the similarity function.

\subsection{Hypercube Ranking}
\label{sec:rank_hyper}
The retrieved documents need to be provided to the LLM as contextual input.
However, long-context RAG methods do not always improve the quality of LLM responses, since they may introduce irrelevant or noisy information \cite{liu2024lost,jin2024long}.
This raises a critical challenge: how to precisely filter and select theme-relevant documents.
To address this, we prioritize returning those documents that fully cover key components derived from the query:
\begin{equation}
    \mathcal{D}_{return} = \{ d_i \in \mathcal{D} \mid \mathcal{E}(q) \subseteq \mathcal{E}(d_i) \}.
\end{equation}

\noindent If no documents fully cover all query components (i.e., $\mathcal{D}_{return}=\emptyset$), we return the next best set of documents with the highest partial coverage:
\begin{equation}
\mathcal{D}^{*}_{\text{return}} = \arg\max_{d_i \in \mathcal{D}} |\mathcal{E}(d_i)|.
\end{equation}

% Taking the example in Figure \ref{fig:hypercube_rag}, \hyperrag prioritizes returning Doc. \texttt{A} and \texttt{B} since they cover all three keywords (see the following table). 
% In cases where Doc. \texttt{A} and \texttt{B} do not exist, then it returns Doc. \texttt{C}, which contains the next best subset of information based on the query decomposition.

% \begin{table}[ht]
% \centering
% \resizebox{0.48\textwidth}{!}{
% \begin{tabular}{l|l}
% \toprule
% \bf Doc.   & \bf Covered Information   \\
% \midrule
% \texttt{A}   & `Melbourne Beach', `Tropical Storm Fay', `Rainfall'    \\
% \texttt{B}   & `Melbourne Beach', `Tropical Storm Fay', `Rainfall'     \\
% \texttt{C}   & `Tropical Storm Fay', `Rainfall'     \\
% \texttt{D}   & `Tropical Storm Fay'     \\
% \bottomrule
% \end{tabular}
% }
% % \caption{Documents.}
% % \label{tab:data_stats}
% \end{table}

% \jh{I am also wondering whether the selectivity (sharpness) should be a concern in ranking since some constants could be much sharper than others.}

\section{Experiments}
\label{sec:exp}
\subsection{Datasets}
\label{sec:data}
% We target our \hyperrag at addressing theme-specific question-answering (QA) tasks on \textit{in-domain} scientific datasets.

To verify the applicability of \hyperrag, three datasets over varying types across different domains are selected: 
(1) \textbf{SciFact} \cite{wadden2020fact}: a dataset of expert-written scientific claims paired with evidence-containing biomedical abstracts annotated with labels and rationales. 
(2) \textbf{LegalBench} \cite{pipitone2024legalbench}: a dedicated QA and retrieval benchmark in the legal domain, entirely human-annotated by legal experts.
(3) \textbf{SciDCC} \cite{mishra2021neuralnere}: a synthesized dataset using LLMs in the geo-science domain, covering ``Geography'', ``Hurricane'', and ``aging dams''. We collected the text corpus from a subset of the Science Daily Climate Change dataset (SciDCC) \cite{mishra2021neuralnere} from the Science Daily website\footnote{\url{https://www.sciencedaily.com/}}. 
``Aging dams'' discusses dam failures in the United States, whose corpus comprises related news articles downloaded from Google News\footnote{\url{https://news.google.com/}}, collected by domain experts. 
The summary is presented in Table \ref{tab:data_stats}.
% and Appendix \ref{sec:qa_type}. 

% \input{tables/data_stat}
\begin{table}[ht]
\centering
\caption{Dataset summary.}
\vspace{-2mm}
\resizebox{0.48\textwidth}{!}{
\begin{tabular}{l|ccccc}
\toprule
\bf Datasets                     & \# Docs   & \# QA   & QA type            & Ques. Length    & Ans. Length    \\
\midrule
\textbf{SciFact} (Medical)       & 5813      & 188     & True or False      & 5$\sim$28       & 1 \\
\textbf{LegalBench} (Law)        & 95        & 977     & Open-Ended         & 16$\sim$39      & 10$\sim$304 \\
\textbf{SciDCC} (Geoscience)     & 1462      & 789     & Open-Ended         & 6$\sim$36       & 10$\sim$150 \\
\bottomrule
\end{tabular}
}
\label{tab:data_stats}
\end{table}

In this work, we focus on studying \hyperrag with one hypercube for each dataset, while more complex domains across various themes may benefit from multiple hypercubes.
We provide a detailed discussion on multiple hypercubes in Appendix \ref{sec:discussion}.

%%%%% QA performance
\begin{table*}[ht!]
\centering
\caption{QA performance ($\%$) on RAG baselines and our Hypercube-RAG.
% (\textcolor{red}{ours}). 
No retrieval means evaluating LLMs with direct inference.
All RAG methods were experimented with the top-5 ($k=5$) retrieved documents using GPT-4o as the base.
The best scores are in \textbf{bold} while the second-best scores are \ul{underlined} (same representations are used in other Tables). 
}
\vspace{-2mm}
\resizebox{0.99\textwidth}{!}{
\begin{tabular}{l|cccc|cccc|cccc}
\toprule
\multirow{3.5}{*}{\textbf{Method}}  & \multicolumn{4}{c|}{\textbf{SciFact (Medical)}}    & \multicolumn{4}{c|}{\textbf{LegalBench (Law)}}   & \multicolumn{4}{c}{\textbf{SciDCC (Geoscience)}} \\
\cmidrule(lr){2-5}  \cmidrule(lr){6-9}  \cmidrule(lr){10-13}
& \multicolumn{2}{c}{\cellcolor{lightyellow}Automatic Metric} & \multicolumn{2}{c|}{\cellcolor{lightblue}LLM-as-Judge} & \multicolumn{2}{c}{\cellcolor{lightyellow}Automatic Metric} & \multicolumn{2}{c}{\cellcolor{lightblue}LLM-as-Judge} & \multicolumn{2}{c}{\cellcolor{lightyellow}Automatic Metric} & \multicolumn{2}{c}{\cellcolor{lightblue}LLM-as-Judge} \\
\cmidrule(lr){2-5}  \cmidrule(lr){6-9}   \cmidrule(lr){10-13}
                        & F1   & Semantic   & Correctness & Completeness           & F1  & Semantic  & Correctness & Completeness        & F1  & Semantic  & Correctness & Completeness\\
\midrule

\rowcolor{gray!10}
\multicolumn{13}{l}{\textbf{\textit{No Retrieval}}} \\
GPT-4o                                      & 81.3  & 84.9 & 81.9  & 81.9           & 22.6  & 56.8 & 52.7 & 32.1           & 26.5  & 64.6    & 59.8 & 29.7 \\
\midrule

\rowcolor{gray!10}
\multicolumn{13}{l}{\textbf{\textit{Sparse and Dense Embedding Retrieval}}} \\
BM25 \cite{robertson1994some}               & 87.8  & 89.8 & 87.7  & 87.7           & 31.8 & 61.2  & 54.6 & 50.9            & \ul{45.3}  & 76.6     & 79.1 & 68.2 \\
Contriever \cite{izacard2021unsupervised}   & 72.3  & 77.1 & 72.3  & 72.3           & 37.8 & 62.4  & 67.1 & 64.3            & 42.6  & 76.5     & 77.2 & 65.6 \\
e5 \cite{wang2024multilingual}              & 90.4  & 92.1 & 90.4  & 90.4           & \ul{38.3} & \ul{63.2}  & \ul{74.1} & \ul{71.5}            & 45.0  & 77.3     & 81.9 & 69.4 \\
NV-Embed-v2 \cite{lee2024nv}                & \ul{91.2}  & \ul{92.6} & \ul{91.3}  & \ul{91.3}           & 32.8 & 60.7  & 54.1 & 50.3            & 44.9  & \ul{77.8}     & 80.4 & \ul{69.8} \\
\midrule

\rowcolor{gray!10}
\multicolumn{13}{l}{\textbf{\textit{Graph-based RAG}}} \\
LightRAG \cite{guo2024lightrag}             & 67.3  & 69.3 & 76.6  & 76.6           & 28.5 & 61.2  & 68.9  & 66.5           & 30.9  & 68.3 & 66.7 & 51.9 \\
GraphRAG \cite{edge2024local}               & 84.9  & 86.2 & 84.8  & 84.8           & 36.3 & 61.8  & 68.4  & 64.7           & 39.2  & 71.4 & 70.4 & 56.6 \\
HippoRAG \cite{gutierrez2024hipporag}       & 85.2  & 87.3 & 86.1  & 86.1           & 34.7 & 60.4  & 67.5  & 63.5                & 41.6  & 72.4 & 75.4 & 59.1 \\
HippoRAG 2 \cite{gutierrez2025rag}          & 88.3  & 90.2 & 89.3  & 89.3           & 37.2 & 62.6  & 70.9  & 69.3                & 43.1  & 75.1 & \bf{84.7} & 58.9 \\
\midrule

\rowcolor{gray!10}
\multicolumn{13}{l}{\textbf{\textit{\textcolor{red}{Our Method}}}} \\
Hypercube-RAG                               & \bf 91.5 & \bf 93.0 & \bf 91.5 & \bf 91.5      & \bf 40.5  & \bf 64.6  & \bf 81.5 & \bf 78.1      & \bf 46.4 & \bf 80.3 & \ul{83.3} & \bf 72.2 \\
\bottomrule

\end{tabular}
}
\label{tab:qa_perform}
\end{table*}

%%%%% Retrieval performance
\begin{table*}[ht!]
\centering
\vspace{+2mm}
\caption{Retrieval performance ($\%$) on RAG baselines and our Hypercube-RAG. All RAG methods were experimented with GPT-4o as the base. GraphRAG and LightRAG were not presented because they do not directly produce passage retrieval results \cite{gutierrez2025rag}. HippoRAG was not tested due to its inferiority compared to HippoRAG 2 \cite{gutierrez2025rag}.
}
\vspace{-2mm}
\resizebox{0.99\textwidth}{!}{
\begin{tabular}{l|cccc|cccc|cccc}
\toprule
\multirow{3.5}{*}{\textbf{Method}}  & \multicolumn{4}{c|}{\textbf{SciFact (Medical)}}     & \multicolumn{4}{c|}{\textbf{LegalBench (Law)}}   & \multicolumn{4}{c}{\textbf{SciDCC (Geoscience)}} \\
\cmidrule(lr){2-5}  \cmidrule(lr){6-9}  \cmidrule(lr){10-13}
& \multicolumn{2}{c}{\cellcolor{lightyellow}$k=3$} & \multicolumn{2}{c|}{\cellcolor{lightblue}$k=5$} & \multicolumn{2}{c}{\cellcolor{lightyellow}$k=3$} & \multicolumn{2}{c}{\cellcolor{lightblue}$k=5$} & \multicolumn{2}{c}{\cellcolor{lightyellow}$k=3$} & \multicolumn{2}{c}{\cellcolor{lightblue}$k=5$} \\
\cmidrule(lr){2-5}  \cmidrule(lr){6-9}   \cmidrule(lr){10-13}
                        & Precision@3  & Recall@3  & Precision@5 & Recall@5     & Precision@3  & Recall@3  & Precision@5 & Recall@5    & Precision@3  & Recall@3  & Precision@5 & Recall@5\\
\midrule

\rowcolor{gray!10}
\multicolumn{13}{l}{\textbf{\textit{Sparse and Dense Embedding Retrieval, and Graph-based RAG}}} \\
BM25 \cite{robertson1994some}               & 27.5 & 75.9 & 18.1  & 81.3        & 12.0  & 36.1  & 7.8  & 38.8      & 28.6  & 86.0     & 17.9  & 89.7 \\
Contriever \cite{izacard2021unsupervised}   & 13.8 & 35.9 & 8.9   & 38.1        & 19.2  & 57.5  & 12.7 & 63.3      & 17.8  & 53.2     & 12.7  & 62.4   \\
e5 \cite{wang2024multilingual}              & 33.0 & 87.7 & 21.2  & 92.8        & \ul{21.7}  & \ul{65.1}  & \ul{13.9} & \ul{69.5}      & \ul{29.0}  & \bf 87.1     & \ul{18.4}  & \ul{91.2} \\
NV-Embed-v2 \cite{lee2024nv}                & 34.4 & 87.9 & \ul{22.9}  & 95.2        & 13.4  & 40.3  & 8.7  & 43.6      & 19.7  & 59.4     & 12.5  & 62.4   \\
% \midrule
% \rowcolor{gray!10}
% \multicolumn{13}{l}{\textbf{\textit{Graph-based RAG}}} \\
% HippoRAG \cite{gutierrez2024hipporag}       & --   & --   & --    & --          & --    & --    & --   & --        & --  & --     & --   & --   \\
HippoRAG 2 \cite{gutierrez2025rag}          & \ul{34.7} & \ul{88.2} & 22.6  & \ul{95.6}        & 12.8  & 38.8  & 12.7 & 64.3      & 18.2  & 57.6     & 12.1   & 61.7   \\
\midrule

\rowcolor{gray!10}
\multicolumn{13}{l}{\textbf{\textit{\textcolor{red}{Our Method}}}} \\
Hypercube-RAG                               & \bf 37.5 & \bf 88.8 & \bf 27.6  & \bf 96.1    & \bf 26.1 & \bf 75.4  & \bf 16.9 & \bf 79.7    & \bf 35.5 & \ul{86.0} & \bf 28.8 & \bf 92.4 \\
\bottomrule

\end{tabular}
}
\label{tab:retrieval_perform}
\end{table*}

\subsection{Baselines}
We select four types of methods for comparison: 1) \textbf{sparse retriever} BM25 \cite{robertson1994some}; 2) \textbf{dense embedding retrieval} methods: Contriever \cite{izacard2022contriever}, e5 \cite{wang2024multilingual}, Nvidia/NV-Embedv2 \cite{lee2024nv}; 3) \textbf{graph-based} methods: GraphRAG \cite{edge2024local}, LightRAG \cite{guo2024lightrag}, and HippoRAG \cite{gutierrez2024hipporag} and HippoRAG 2 \cite{gutierrez2025rag}; and (4) \textbf{LLMs without retrieval}. 
We implement the sparse BM25 retriever using a GitHub repository\footnote{\url{https://github.com/dorianbrown/rank_bm25}}.
For dense embedding retrievers, we utilize their models uploaded to Hugging Face.
Experiments for graph-based baselines using their official GitHub repositories. 
All experiments were conducted on a single A100 GPU with 80 GB memory.

% \begin{figure*}[ht]
% \centering
% \begin{subfigure}[t]{0.45\textwidth}
%   \centering
%   \includegraphics[width=0.99\textwidth]{figures/bar_plot_hurricane.jpg}
%   \caption{Hurricane}
% \end{subfigure}
% \hfill
% \begin{subfigure}[t]{0.45\textwidth}
%   \centering
%   \includegraphics[width=0.99\textwidth]{figures/bar_plot_geography.jpg}
%   \caption{Geography}
% \end{subfigure}
% \caption{Performance comparison with various LLMs (only the range between 50\% and 90\% shown for clarity).}
% \label{fig:bar_plot}
% \end{figure*}

\subsection{Evaluation Metrics}
We utilize multiple metrics for a comprehensive evaluation of different methods.
To measure the QA performance, following the work \cite{cui2025timer}, we evaluate the quality of LLMs' responses using automated metrics derived from token-level representations, including F1 and Semantic score \cite{chandrasekaran2021evolution} to provide the standard assessment of response quality. 
Since the settings are open-text responses, we also employ LLM-as-a-Judge that assesses correctness and completeness. 
All LLM-based evaluations use GPT-4o as the judge. The prompts and other parameters used are in Appendix \ref{sec:prompt}.
To measure the QA performance, following the HippoRAG work \cite{gutierrez2025rag}, precison@$k$ and recall@$k$ rates are computed to measure the retrieval performance.

\section{Results}
\label{sec:result}
%
% We show and analyze the experimental results for accuracy, efficiency, and explainability.
\subsection{Accuracy}
% Tables \ref{tab:qa_perform} and \ref{tab:retrieval_perform} report the results for QA and retrieval performance, respectively. We present the analysis as follows. 

\subsubsection{QA performance.}
Table \ref{tab:qa_perform} reports the results for QA performance. The observations and findings are as follows. 
We noticed that the state-of-the-art baselines present varying performance over different datasets. Most of them perform better on SciFact than the other two datasets. 
Notably, \hyperrag consistently outperforms other baselines across all three datasets, demonstrating its effectiveness in enhancing the capabilities of LLMs for scientific question-answering tasks. 
\hyperrag surpasses the second-best models with an average improvement by $3.7\%$ on F1 score, $3.3\%$ on semantic score, $5.4\%$ on correctness score, and $3.7\%$ on completeness score, evaluated across three datasets.
The most significant improvements were observed on LLM-as-judge scores over the LegalBench and SciDCC datasets.
We also observed that all RAG methods perform much better than direct inference of LLMs without retrieval, as expected, showing the effectiveness of incorporating external knowledge for \emph{scientific} QA tasks.
% Finally, we also evaluate \hyperrag against a range of baseline methods across multiple LLMs developed by different companies. As illustrated in Figure~\ref{fig:bar_plot}, our method achieves superior performance across all models regardless of the underlying LLM architectures.

\subsubsection{Retrieval performance.}
Table \ref{tab:retrieval_perform} reports retrieval results for datasets with supporting passage annotations and models that explicitly retrieve passages. 
The retrieval accuracy of baseline methods also varied by dataset: HippoRAG 2 was the second-best model on SciFact, while e5 achieved this rank on both LegalBench and SciDCC.
Compared to the strongest baselines, \hyperrag exhibits superior precision and recall scores with an average improvement by $3.9\%$ and $0.6\%$ on SciFact, $3.7\%$ and $10.2\%$ on LegalBench, and $8.5\%$ and $0.1\%$ on SciDCC, respectively. 
These results highlight \hyperrag as a state-of-the-art RAG system that enhances both retrieval and QA performance.

\subsection{Efficiency}
We study the response speed of diverse RAG methods over the corpus size and report the retrieval time for one query in Table \ref{tab:efficiency}. 
% We break down our analysis as follows.

\subsubsection{Within corpus size $k$.}
BM25 achieves the fastest retrieval among all evaluated methods without surprise. 
In contrast, semantic and graph-based RAGs incur significantly higher retrieval costs.
This is particularly pronounced for graph-based methods due to the computational burden of search paths in large-scale graph structures, posing scalability challenges. 
Notably, our \hyperrag substantially reduces the retrieval time by \textbf{one to two orders of magnitude} compared to both semantic and graph-based methods, underscoring the effectiveness of the hypercube structure in optimizing retrieval efficiency. 
We attribute this to the retrieval with \emph{compact, multi-dimensional cube labels} introduced in Section \ref{sec:retrieve_hyper}. 
Although it is slightly slower than BM25, the marginal increase in retrieval time is a worthwhile trade-off for significantly improved retrieval accuracy and response quality, as shown in Tables \ref{tab:qa_perform} and \ref{tab:retrieval_perform}.

\begin{table}[ht!]
\centering
\caption{Retrieval time ($ms$) vs.\ corpus size ($k$), experimenting on the Hurricane-related data, a subset of SciDCC (LLM base: GPT-4o).
The retrieval time with $^\dag$ performs a spike for some reasons, though the experiments were repeated three times.
}
\vspace{-2mm}
\resizebox{0.47\textwidth}{!}{
\begin{tabular}{l|cccc|c}
\toprule
\bf Methods   &\bf $k/8$  &\bf $k/4$  &\bf $k/2$  &\bf $k$  &\bf $14k$\\
\midrule
BM25                    & 0.5    & 0.9      & 1.6      & 3.2      & 51.1 \\
% Contriever              & 12.6   & 16.4     & 28.6     & 65.4     & XXX \\
e5                      & 14.4   & 17.4     & 19.1     & 37.8     & 82.4 \\
GraphRAG                & 114.7  & 348.2    & 1179.4   & 5260.6   & 45135.7 \\
% HippoRAG 2              & 1274.02 & 1734.81   & 1820.60   & 2108.71  \\  % averag all datasets
HippoRAG 2              & 1575.0$^\dag$ & 850.0    & 1722.8   & 2403.6   & 7490.4 \\      % first question
% \midrule
Hypercube-RAG           & 0.7    & 1.5      & 3.3      & 7.1      & 22.1\\
\bottomrule
\end{tabular}
}
\label{tab:efficiency}
\end{table}

\subsubsection{Beyond corpus size $k$.}
To mimic the real-world scenario that the external knowledge database is usually large and noisy, we expand the hurricane-related corpus from 844 to 11,539 documents (around 14 times larger).
The added documents are related to other topics, such as ``Pollution'' or ``Ozone Holes'', introducing additional irrelevant documents (``noise'').
\hyperrag maintains the lowest retrieval time per query (see the last column in Table \ref{tab:efficiency}), highlighting the efficiency of the cube label-based retrieval mechanism. 
This is likely because most noisy documents result in empty or sparsely populated cube cells, as they lack corresponding \emph{fine-grained} labels pre-defined in our hypercube. As a result, despite a significantly expanded corpus, the hypercube can efficiently bypass these irrelevant documents.
In contrast, other baseline methods exhibit greater sensitivity to corpus size, resulting in increased retrieval times.
In addition, Table \ref{tab:scale} shows that the accuracy of the evaluated methods drops due to the inclusion of these noisy off-topic documents. However, \hyperrag still outperforms others, demonstrating robust resilience to noise.
\begin{table}[ht]
\centering
\caption{Accuracy vs.\ Corpus Size. $\downarrow$ represents the accuracy dropped with the corpus size from $k$ to $14k$.}
\vspace{-2mm}
\resizebox{0.47\textwidth}{!}{
\begin{tabular}{l|cc|c}
\toprule
\bf Methods      & Semantic Score with $k$  & Semantic Score with $14k$    & $\Delta$ Score ($\downarrow$)\\
\midrule
BM25            & 77.5  & 77.2            & 0.3  \\
e5              & 79.1  & 76.3            & 2.8  \\
GraphRAG        & 72.8  & 70.5            & 2.3  \\
HippoRAG 2      & 76.3  & 73.7            & 2.6  \\
Hypercube-RAG   & 82.5  & 80.3            & 1.9  \\
\bottomrule
\end{tabular}
}
\label{tab:scale}
\end{table}

\subsection{Explainability}
The retrieval process with our hypercube is inherently explainable through its associated labels in the cube cell(s). 
Considering the query \textit{``How much rainfall did Melbourne Beach, Florida receive from Tropical Storm Fay?''} in Figure \ref{fig:hypercube_rag}, \hyperrag returns relevant documents by matching the decomposed components with fine-grained labels, ``Melbourne Beach'', ``Tropical Storm Fay'', and ``rainfall'', along hypercube dimensions: \textit{Location}, \textit{Event}, and \textit{Theme}. 
Table \ref{tab:explainability} shows that three retrieved documents (Doc. \texttt{565}, \texttt{246}, \texttt{535}), which are represented with fine-grained cube labels.
These cube labels clearly explain why those documents are returned.
For example, Doc. \texttt{565} is ranked the first to return, since it contains all three key components of the query. The detailed document contents are included in Table \ref{tab:doc_content}.

\begin{table}[ht]
\centering
\vspace{+2mm}
\caption{Documents represented in the hypercube.}
\vspace{-2mm}
\resizebox{0.47\textwidth}{!}{
\begin{tabular}{l|ccc}
\toprule
\bf Doc.       & \bf Location             & \bf Event                   & \bf Theme   \\
\midrule
\texttt{565}   & `Melbourne Beach': 1  & `Tropical Storm Fay': 1   & `Rain': 5    \\
\texttt{246}   & `Florida': 1          & `Tropical Storm Fay': 1   & --    \\
\texttt{535}   & `Florida': 1          & --                        & -- \\
\bottomrule
\end{tabular}
}
\label{tab:explainability}
\end{table}

\begin{table*}[ht!]
\small
\caption{Document ID and content in the corpus. Colored words and phrases are used for retrieval explainability in Table \ref{tab:explainability}.}
\vspace{-2mm}
\begin{tabular}{p{0.14\textwidth}|p{0.8\textwidth}}
\toprule
\textbf{Document ID} & \textbf{Document Content} \\
\midrule
\parbox[t]{0.14\textwidth}{565} & 
\parbox[t]{0.8\textwidth}{\ldots  The 2008 Atlantic hurricane season started early with the formation of Tropical Storm Arthur on May 30, from the remnants of the eastern Pacific Ocean’s first storm, Alma, which
crossed Central America and reformed in the Gulf of Mexico. It took one month and four days
for the next storm to form, Bertha. Once a powerful Category 3 hurricane, now a tropical depression, Gustav moved from northwest Louisiana into northeastern Texas and into Arkansas by Sept. 3. Like \textcolor{red!90}{Tropical Storm Fay} in August, Gustav's legacy will lie in large \textcolor{yellow!90}{rainfall} totals. According to the National Hurricane Center discussion on Sept. 2, "Storm total \textcolor{yellow!90}{rainfalls} are expected to be five to ten Inches with isolated maximums of 15 inches over portions of Louisiana, Arkansas and Mississippi. \textcolor{yellow!90}{Rainfall} amounts of 4-8 inches have been already reported in parts of Alabama, Mississippi and Louisiana. \ldots 
% On Sept. 4, Ike strengthened into a major hurricane, Category 4 on the Saffir-Simpson Scale, with maximum sustained winds near 145 mph. Ike is forecast to head west and may also affect the Bahamas. Behind Ike, on Sept. 2, the tenth tropical depression in the Atlantic Ocean basin was born. Tropical Depression 10 formed west of the African coast, so it has a long way to go before it has any impact on the U.S. or the Caribbean. 
By the late morning, that tropical cyclone became Tropical Storm Josephine. In August, \textcolor{red!90}{Fay}'s ten-day romp from the U.S. Southeast northward up the Appalachian Mountains seemed like a harbinger for September's storms. \textcolor{red!90}{Fay} took her time going northward and dumped tremendous amounts of \textcolor{yellow!90}{rain} along the way. \textcolor{blue}{Melbourne Beach, Florida}, received as much as 25.28 inches of \textcolor{yellow!90}{rain}. Other cities in various states reported high totals: Thomasville, Ga., reported 17.43 inches; Camden, Ala., received 6.85 inches; Beaufort, S.C., received 6.11 inches; Carthage, Tenn., reported 5.30 inches, and Charlotte, N.C., reported 5.90 inches. \textcolor{red!90}{Fay} was a perfect example of how weaker tropical storms can cause flooding inland \ldots} \\
\midrule
\parbox[t]{0.14\textwidth}{246} & 
\parbox[t]{0.8\textwidth}{\ldots Among the already apparent evidence: Dunes that historically protected Kennedy Space Center from high seas even during the worst storms were leveled during \textcolor{red!90}{Tropical Storm Fay} in 2008, Hurricane Irene in 2011 and Hurricane Sandy in 2012. A stretch of beachfront railroad track built by NASA in the early 1960s that runs parallel to the shoreline has been topped by waves repeatedly during recent storms. \ldots The research came about after NASA partnered with the U.S. Geological Survey and UF to figure out why chronic erosion was happening along a roughly 6-mile stretch of beach between launch pads 39A and 39B -- the ones used for Space Shuttle and Apollo missions. The problem had been occurring for years but seemed to be growing worse, beginning with the spate of hurricanes that struck \textcolor{blue}{Florida} in 2004. Jaeger said he, Adams and doctoral students Shaun Kline and Rich Mackenzie determined the cause was a gap in a near-shore sandbar that funnels the sea toward that section of beach. \ldots} \\
\midrule
\parbox[t]{0.14\textwidth}{535} & 
\parbox[t]{0.8\textwidth}{\ldots 
% Our theory would suggest that seepage caused by underwater flow will continue to erode and weaken the levee system around New Orleans, but the rate of this erosion should gradually slow with time, says Straub. 
Hopefully this research will aid the U.S. Army Corps of Engineers in identifying levees that need repair and assessing the lifespan of structures like the MRGO that are not planned for upkeep.
Using fieldwork conducted in the \textcolor{blue}{Florida Panhandle}, Straub and his fellow researchers were able to better understand the process of seepage erosion, which occurs when the re-emergence of groundwater at the surface shapes the Earth's topography. The paper suggests that the velocity at which channel heads advance is proportional to the flux of groundwater to the heads. The researchers used field observations and numerical modeling to come up with the theory. \ldots} \\
\bottomrule
\end{tabular}
\label{tab:doc_content}
\end{table*}

\begin{figure*}[ht!]
\vspace{+2mm}
\makebox[\textwidth][c]{%
\begin{tcolorbox}[colback=white,colframe=viridis3!75,title={\bf Case Study 2}, width=\textwidth] %viridis3

\footnotesize
\textbf{Query: \textit{What is the likelihood of an above-normal, near-normal and below-normal hurricane season at the Atlantic, according to the Climate Prediction Center?}} 
\begin{center}
    \vspace{-1em}
    \begin{tikzpicture}
      \draw[dashed] (0,0) -- (\linewidth,0)
      node[midway, above]{\small \textbf{Hypercube-RAG}};
    \end{tikzpicture}
\end{center}
\footnotesize
\textbf{> Answer}: \textit{Above-normal: 85\%, Near-normal: 10\%, Below-normal: 5\%}  ({\large \textcolor{blue!90}{\bf \checkmark}}) \\
\texttt{Retrieved Docs}: [19, 230] \\
\texttt{Doc 19:} \textit{\dots Based on the ACE projection, combined with the above-average numbers of named storms and hurricanes, the likelihood of an above-normal \textcolor{blue}{Atlantic hurricane season} has increased to \textcolor{yellow!90}{85\%}, with only a \textcolor{yellow!90}{10\%} chance of a near-normal season and a \textcolor{yellow!90}{5\%} chance of a below-normal season \dots, said Gerry Bell, Ph.D., lead seasonal hurricane forecaster at NOAA's \textcolor{red!80}{Climate Prediction Center} \dots} 
\begin{center}
    \vspace{-1em}
    \begin{tikzpicture}
      \draw[dashed] (0,0) -- (\linewidth,0)
      node[midway, above]{\small \textbf{Sparse BM25}};
    \end{tikzpicture}
\end{center}
\footnotesize
\textbf{> Answer}: \textit{10\% above-normal, 20\% near-normal, 70\% below-normal.} ( {\textcolor{red}{\ding{55}}} ) \\
\texttt{Retrieved Docs}: [230, 273] \\
\texttt{Doc 230:} \textit{\dots NOAA is predicting a 70 percent likelihood of 6 to 11 named storms, of which 3 to 6 could become hurricanes \dots While a below-normal season is likely (\textcolor{yellow!90}{70 percent}), there is also a \textcolor{yellow!90}{20 percent} chance of a near-normal season, and a \textcolor{yellow!90}{10 percent} chance of an above-normal season \dots} 
\begin{center}
    \vspace{-1em}
    \begin{tikzpicture}
      \draw[dashed] (0,0) -- (\linewidth,0)
      node[midway, above]{\small \textbf{Semantic-RAG (e5)}};
    \end{tikzpicture}
\end{center}
\footnotesize
\textbf{> Answer}: \textit{Above-normal: 10\%, Near-normal: 20\%, Below-normal: 70\%.} ( {\textcolor{red}{\ding{55}}} ) \\
\texttt{Retrieved Docs}: [230, 636] \\
\texttt{Doc 230:} \textit{\dots NOAA is predicting a 70 percent likelihood of 6 to 11 named storms, of which 3 to 6 could become hurricanes \dots While a below-normal season is likely (\textcolor{yellow!90}{70 percent}), there is also a \textcolor{yellow!90}{20 percent} chance of a near-normal season, and a \textcolor{yellow!90}{10 percent} chance of an above-normal season \dots} 
\begin{center}
    \vspace{-1em}
    \begin{tikzpicture}
      \draw[dashed] (0,0) -- (\linewidth,0)
      node[midway, above]{\small \textbf{Graph-based HippoRAG-2}};
    \end{tikzpicture}
\end{center}
% \vspace{+0.5em}
\footnotesize
\textbf{> Answer}: \textit{65\%, 25\%, 10\% chance for above-normal season, near-normal season, below-normal season.} ( {\textcolor{red}{\ding{55}}} ) \\
\texttt{Retrieved Docs}: [583, 275] \\  %583, 275, 575, 386, 273
\texttt{Doc 583:} \textit{\dots The Climate Prediction Center outlook calls for considerable activity with a \textcolor{yellow!90}{65 percent} probability of an above normal season and a \textcolor{yellow!90}{25 percent} probability of a near normal season. This means there is a 90 percent chance of a near or above normal season. \dots}

\end{tcolorbox}
}
\vspace{-3mm}
\caption{Comparison of different RAG methods on the same query.}
\label{fig:case_study2}
\end{figure*}

\subsection{Ablation Study}
\begin{figure}[H]
\centering
  \includegraphics[width=0.99\columnwidth]{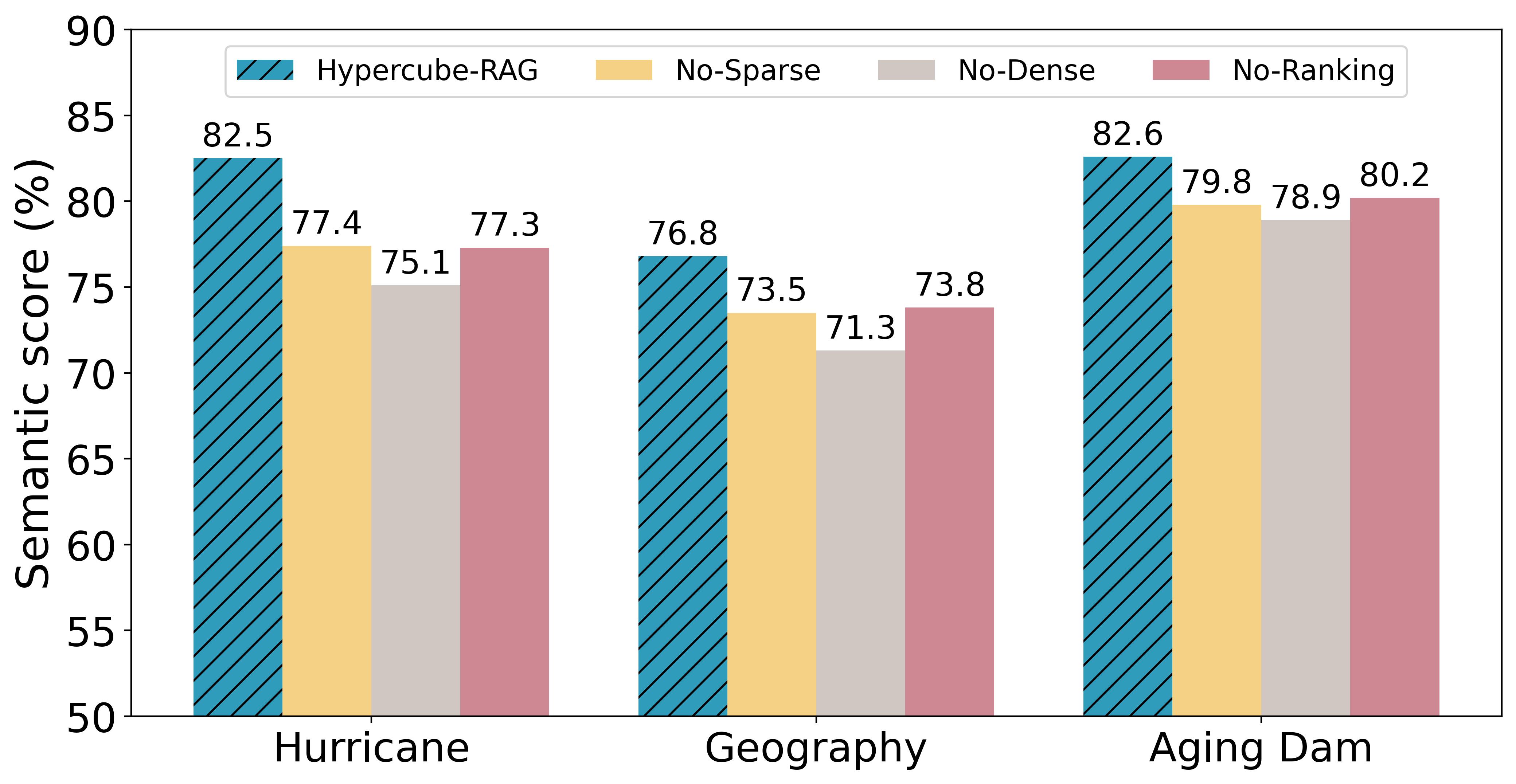}
  \vspace{-4mm}
  \caption{Ablation study on model components.} 
  \label{fig:ablation}
\end{figure}

\textbf{Model Components.}
As shown in Section \ref{sec:retrieve_hyper}, \hyperrag combines two retrieval strategies: sparse exact matching and dense embedding retrieval. Section \ref{sec:rank_hyper} introduces the document ranking algorithm used to filter and prioritize the retrieved results. 
To study their effectiveness, we select the SciDCC dataset and conduct an ablation study by removing each of them from \hyperrag, which are represented as No-Sparse, No-Dense, and No-Ranking, respectively. 
Figure \ref{fig:ablation} shows that the full version of \hyperrag consistently outperforms all other variants, verifying positive contributions of each constituent. 
More specifically, it justifies the benefit of coupling these two strategies for scientific question-answering tasks, and the necessity of the ranking algorithm to filter possible ``noisy'' documents.
% The ablation study on hypercube dimensions is in Appendix \ref{sec:more_ablation}.
%

\noindent \textbf{Hypercube Dimensions.} We also conduct an ablation study on the hypercube dimensions: \textit{Location}, \textit{Event}, \textit{Date}, \textit{Organization}, \textit{Person}, and \textit{Theme}. 
To study their effectiveness, we remove each of the dimensions, which are represented as \textit{No-Location}, \textit{No-Event}, \textit{No-Date}, \textit{No-Organization}, \textit{No-Person}, and \textit{No-Theme}, respectively. 
Table \ref{tab:abs_dim} shows that the full version of \hyperrag consistently outperforms all other variants, verifying positive contributions of each dimension. 
We observe that the performance drops the most when removing the \textit{Location} and \textit{Theme} dimensions, justifying their high importance for theme-specific QA tasks in geoscience.
We claim that these dimensions in a hypercube could be changed over datasets with different focuses of interest. The hypercube dimensions used for other datasets are provided in Appendix \ref{sec:hypercube_design}.

\begin{table}[ht!]
\centering
\caption{Ablation study on hypercube dimensions.}
% for the SciDCC dataset. The reported values are semantic scores ($\%$).}
\vspace{-2mm}
\resizebox{0.48\textwidth}{!}{
\begin{tabular}{l|ccc}
\toprule
\bf Dimension           & \bf Hurricane & \bf Geography & \bf Aging Dam \\
\midrule
No-Location             & 78.7          & 65.9          & 77.5         \\
No-Event                & 79.6          & 75.9          & 80.0        \\
No-Date                 & 79.7          & 74.5          & 80.4        \\
No-Organization         & 79.6          & 74.6          & 80.2       \\
No-Person               & 80.2          & 73.2          & 81.4         \\
No-Theme                & 77.9          & 71.7          & 81.1          \\
\midrule
FULL                    & 82.5          & 76.8          & 82.6          \\
\bottomrule
\end{tabular}
}
\label{tab:abs_dim}
\end{table}

\subsection{Parameter Study}
\begin{figure}[t]
\centering
  \includegraphics[width=0.99\columnwidth]{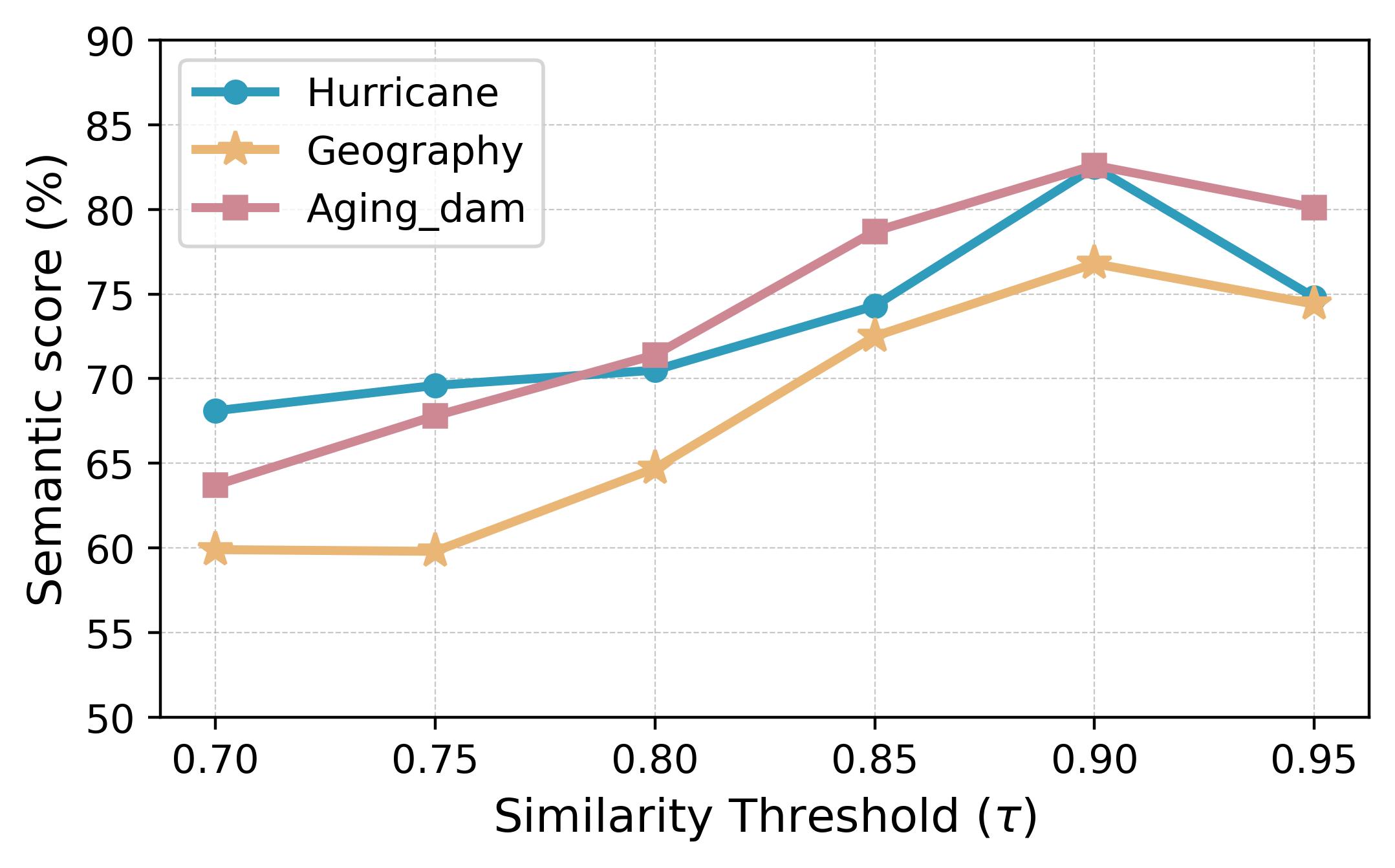}
  \vspace{-2mm}
  \caption{Performance vs.\ similarity threshold.} 
  \label{fig:threshold}
  \vspace{-3mm}
\end{figure}

\hyperrag combines sparse exact match and semantic retrieval strategies, as introduced in Section \ref{sec:retrieve_hyper}. 
We investigate the effect of the semantic retrieval strategy on the quality of LLM responses under varying similarity thresholds, denoted by $\tau$. 
Figure~\ref{fig:threshold} shows that performance improves with increasing threshold values when $\tau \leq 0.9$, suggesting that higher similarity scores are more effective for retrieving semantically relevant documents.
However, when $\tau > 0.9$, the behavior of semantic RAG begins to gradually degrade to the sparse exact matching strategy, resulting in a decline in performance.
This is primarily due to overly strict matching criteria, which reduce the number of retrieved documents and thus limit the available context.

\subsection{Case Study}
An additional case study in Figure \ref{fig:case_study2} 
presents that Sparse BM25, Semantic-RAG (e5), and HippoRAG 2 fail to generate the correct answer due to inaccurate document retrieval. 
These methods tend to prioritize returning documents that are semantically similar in the embedding space.
For example, semantic embedding methods tend to retrieve those documents that largely mention \textit{above-normal, near-normal, and below-normal hurricane season}. 
However, these common terminologies may overshadow more specific, less frequent, but critical information, such as the location ``Atlantic'' and the organization ``Climate Prediction Center''. 
In contrast, our \hyperrag returns the correct answer.
%, which can be attributed to its precise and comprehensive retrieval processes. 
It accurately identifies and retrieves documents containing key elements of the query, including ``Atlantic'', ``hurricane season'', and ``Climate Prediction Center'', corresponding to the hypercube dimensions, \texttt{LOCATON}, \texttt{THEME}, and \texttt{ORGANIZATION}, respectively.
This retrieval approach intuitively prioritizes specific contextual information over abstract, ambiguous references such as \textit{above-, near-, and below-normal} alone.

\section{Discussion and Conclusions}
\label{sec:conclude}
In this work, we start by analyzing the strengths and limitations of existing RAG methods for scientific question-answering, which tend to perform poorly in either accuracy, efficiency, or explainability of the retrieval process.
To retain their advantages while addressing key challenges, we introduce \texttt{Hypercube}, a novel structure representing knowledge in a compact, multi-dimensional space.
Cube cell(s) are associated with compact, fine-grained labels extracted from the documents using LLMs.
Built on the hypercube, \hyperrag is proposed to innovate the knowledge retrieval based on those cube labels.
The cube label-based retrieval integrates sparse lexical and dense semantic strategies, making it accurate, efficient, and inherently explainable.
We conduct experiments on three datasets across different domains comprehensively, and report the results using four retrieval metrics and two QA metrics, demonstrating that our method consistently outperforms existing baselines in both retrieval and question-answering accuracy.
Additionally, we reveal a higher retrieval efficiency of \hyperrag than existing approaches, and verify the explainable retrieval process.
Overall, we conclude that \hyperrag achieves high accuracy and efficiency, while providing interpretable provenance for retrieval.

% % The acknowledgments section is defined using the "acks" environment
% % (and NOT an unnumbered section). This ensures the proper
% % identification of the section in the article metadata, and the
% % consistent spelling of the heading.
\begin{acks}
This work is supported by the Institute for Geospatial Understanding through an Integrative Discovery Environment (I-GUIDE), which is funded by the National Science Foundation (NSF) under award number: 2118329. Any opinions, findings, conclusions, or recommendations expressed herein are those of the authors and do not necessarily represent the views of NSF.

\end{acks}

% \newpage
%% The next two lines define the bibliography style to be used, and the bibliography file.
\bibliographystyle{ACM-Reference-Format}
\bibliography{reference}

\newpage
\appendix
\onecolumn

\section*{Appendix} 
\label{sec:appendix}
% =================== Clustering for Hypercube Design ===================
\section{Hypercube}
\label{sec:hypercube_design}
\subsection{Hypercube Design and Construction}
Comprehensively understanding the corpus is critical for pre-defining hypercube dimensions, because these dimensions structurally represent the knowledge (i.e., documents) in the multi-dimensional space. We follow these steps to determine the hypercube dimensions from the corpus.
\tikzmarknode[mycircled]{a1}{1} \textbf{Entity extraction:} We employ pre-trained language models to extract entities from the entire corpus. These entities are saved in the entity pool without categorization.
\tikzmarknode[mycircled]{a1}{2} \textbf{Entity clustering:} The K-means clustering algorithm is applied to group these entities into semantically coherent clusters.
For example, countries, states, provinces, and specific places are clustered together within a location cluster, while minutes, hours, days, weeks, months, and years are in the date/time cluster. Similarly, the entities with the semantic meaning should be grouped in one cluster, such as those different medicine names, different hurricane names, different law agreements, etc.
\tikzmarknode[mycircled]{a1}{3} \textbf{Cluster summarization:} From step 2, we do not know what clusters or categories the entities belong to.
Therefore, we let a large language model (LLM) generate concise summaries that characterize the underlying cluster categories or dimensions.
\tikzmarknode[mycircled]{a1}{5} \textbf{Clusters to Dimensions:} 
Since some entities can be overlapped in multiple clusters, we use LLMs to consolidate and shrink those summarized cluster names by merging those overlaps, and finalize a small number of high-level clusters as the hypercube dimensions. This dimension-identification process for hypercube design has demonstrated strong alignment with domain expert expectations.
\tikzmarknode[mycircled]{a1}{5} \textbf{Document indexing:} Once the dimensions are pre-defined, we leverage LLMs to extract entities along each dimension, and index documents in a multi-dimensional space.

\begin{figure*}[ht]
\centering
  \includegraphics[width=0.98\columnwidth]{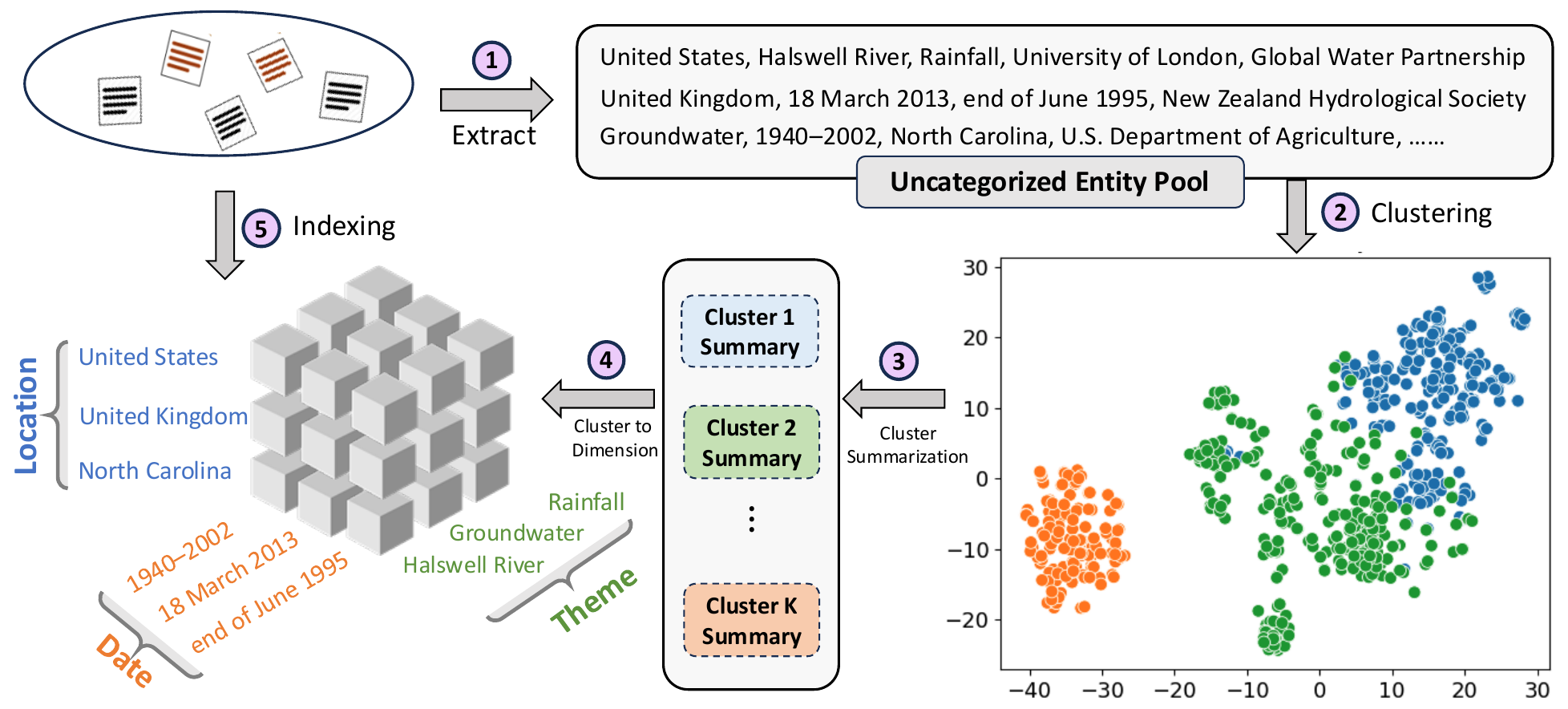}
  \vspace{-2mm}
  \caption{Hypercube Structure Design and Construction.} 
  \label{fig:clustering}
\end{figure*}

\subsection{Hypercube Structure (Dimensions)}
Table \ref{tab:hypercube_dim} presents the hypercube dimensions used for different datasets.
\begin{table}[ht!]
\small
\caption{Document ID and content in the corpus. Colored words and phrases are used for retrieval explainability in Table \ref{tab:explainability}.}
\vspace{-2mm}
\begin{tabular}{p{0.18\textwidth}|p{0.65\textwidth}}
\toprule
\textbf{Datasets} & \textbf{Hypercube Dimensions} \\
\midrule
\parbox[t]{0.18\textwidth}{SciFact} & 
\parbox[t]{0.65\textwidth}{``date'', `location'', ``person'', ``quantity'', ``organizations\_research\_initiatives'', \\ ``medicine\_health'', ``genetics\_biology'', ``immunology\_neuroscience'', ``pharmacology''} \\
\midrule
\parbox[t]{0.18\textwidth}{LegalBench} & 
\parbox[t]{0.65\textwidth}{``date'', ``organization'', ``quantity'', ``location'', ``person'',  ``company'', ``money\_finance'', ``relationship'', \\ ``law\_agreement\_regulation'', ``agreement\_information''} \\
\midrule
\parbox[t]{0.18\textwidth}{SciDCC (Hurricane)} & 
\parbox[t]{0.65\textwidth}{``location'', ``person'', ``event'', ``organization'', ``theme'', ``date''} \\
\midrule
\parbox[t]{0.18\textwidth}{SciDCC (Geography)} & 
\parbox[t]{0.65\textwidth}{``location'', ``person'', ``event'', ``organization'', ``theme'', ``date''} \\
\midrule
\parbox[t]{0.18\textwidth}{SciDCC (Aging\_Dam)} & 
\parbox[t]{0.65\textwidth}{``dam'', ``date'', ``facility'', ``location'', ``organization'', ``person'', ``event'', ``quant'', 'watershed'} \\
\bottomrule
\end{tabular}
\label{tab:hypercube_dim}
\end{table}

% =================== Prompt Templates ===================
\section{Prompt Templates}
\label{sec:prompt}
% ======================== Prompt template for the query decomposition ========================
% \begin{figure}[ht!]
\makebox[0.95\textwidth][c]{%
\begin{tcolorbox}[colback=white,colframe=viridis3!80,title={\bf Prompt template for the query decomposition}, width=0.95\textwidth] %

\footnotesize
You are an expert on question understanding. Your task is to: \\

1. Comprehend the given question: understand what the question asks, how to answer it step by step, and all concepts, aspects, or directions that are relevant to each step.

2. Compose queries to retrieve documents for answering the question: each document is indexed by the entities or phrases occurred inside and those entities or phrases lie within the following dimensions: \textcolor{red!80}{\{dimensions\}}.  \\

For each of the above dimensions, synthesize queries that are informative, self-complete, and mostly likely to retrieve target documents for answering the question. Note that each of your queries should be an entity or a short phrase and its associated dimension. \\

Example Input: ``How much rainfall did Melbourne Beach, Florida receive from Tropical Storm Fay?'' \\ 

Example Output:

1. query\_dimension: `location'; query\_content: `Atlantic';

2. query\_dimension: `event'; query\_content: `Tropical Storm Fay';

3. query\_dimension: 'theme'; query\_content: `rainfall'

\end{tcolorbox}
}
% ======================== Prompt template for question-answering using LLMs ========================
% \begin{figure}[ht!]
\vspace{+4mm}

\noindent
\makebox[0.95\textwidth][c]{%
\begin{tcolorbox}[colback=white,colframe=red!50,title={\bf Prompt template for Question-Answering}, width=0.95\textwidth] 

\footnotesize
Answer the question based on the given retrieved documents. \\

Question: \textcolor{red!70}{\{question\}}, \\

Retrieved documents: \textcolor{red!70}{\{retrieved document\}}, \\ 

Output requirements: \\
 - If the query asks the quantitative analysis, such as starting with "How many", "How much", "How much greater", "How wide", "What percentage", "How far", "How long", "How old", "What portion", "What depth", please directly output the quantitative answer as short as possible.
 
- If the query asks specific information, such as "what percentage", "what likelihood", "which years", "who", please directly output the answer without explanation or other information.
\end{tcolorbox}
}
% \end{figure}
\vspace{+4mm}

\noindent
% ======================== Prompt template for LLM-as-judge ========================
% \begin{figure}[ht!]
\makebox[0.95\textwidth][c]{%
\begin{tcolorbox}[colback=white,colframe=gray!85,title={\bf Prompt template for LLM-as-a-judge}, width=0.95\textwidth] %
\footnotesize
% You are an expert evaluator on natural sciences, such as hurricanes, geography, and aging dams. 
Your task is to assess the predicted answer generated by AI models compared to the gold (reference) answer. Please evaluate the predicted answer on correctness and completeness. \\

Question: \textcolor{red!70}{\{question\}}, \\
Gold Answer: \textcolor{red!70}{\{gold\_answer\}}, \\
Predicted Answer: \textcolor{red!70}{\{predicted\_answer\}}. \\

Evaluate criteria: 

1. Correctness (0 or 1):

- Score 1 if the predicted answer is generally accurate and aligns with the key points in the reference answer.

- Score 0 if it has factual errors or misrepresents key information.

2. Completeness (0 or 1):

- Score 1 if the predicted answer covers the main points present in the reference answer.

- Score 0 if it misses essential information or fails to address the core of the question. \\

Output your evaluation in the following JSON format:
\{``correctness'': int, ``completeness'': int, ``explanation'': your brief explanation\}
\end{tcolorbox}
}
% \label{fig:llm_as_judge}
% \end{figure}
\vspace{+4mm}

\noindent
% \begin{figure}[ht!]
\makebox[0.95\textwidth][c]{%
\begin{tcolorbox}[colback=white,colframe=yellow!60,title={\bf Prompt to extract entities/terms of LOCATION}, width=0.95\textwidth] %viridis3
\footnotesize
    Please extract countries, states, provinces, cities, towns, and specific addresses from the given document, and save them in the following format ...
\end{tcolorbox}
}
% \label{fig:llm_as_judge}
% \end{figure}

% =================== Discussion ===================
\newpage
\section{Discussion}
\label{sec:discussion}
In the following, we discuss different scenarios of hypercube usage: simple query, long query with multiple themes, and diverse query with shifted themes.
% \subsection{Why does \hyperrag outperform other RAG methods?}
% Our \hyperrag is particularly proposed for domain-focused analysis, where domain-specific information, e.g., location, date, event, event, theme, is central to the query. 
% In such cases with theme-specific information, our hypercube enables precise and efficient retrieval by accessing \textbf{only one or a few cells in a theme-specific hypercube}, as each cell encapsulates multiple dimensions of information associated with a document.
% These fine-grained document labels can be represented in a hypercube along different dimensions.
% When a query comes in, accessing one or a few hypercubes presents a higher chance of retrieving highly relevant documents covering all key information contained in a query - \emph{Accuracy}.
% Additionally, retrieval is operated with the cube labels significantly speeds up the retrieval time since those labels are in a compact space - \emph{Efficiency}. In contrast, sparse exact match and dense embedding retrieval require traversing all documents and contents inside.
% Moreover, the retrieval process is inherently explainable, as the searching is based on the document's labels represented by a hypercube - \emph{Explainability}.

\subsection{Case 1: simple query}
\textit{How much rainfall did Melbourne Beach,
Florida receive from Tropical Storm Fay?} The relevant documents can be retrieved from one cube cell in a hypercube <\textcolor{black}{Melbourne Beach}, \textcolor{black}{rainfall}, \textcolor{black}{Tropical Storm Fay} >.
% <\textcolor{blue}{Melbourne Beach}, \textcolor{yellow!90}{rainfall}, \textcolor{red!90}{Tropical Storm Fay} >.
%
\begin{figure}[H]
\centering  \includegraphics[width=0.5\columnwidth]{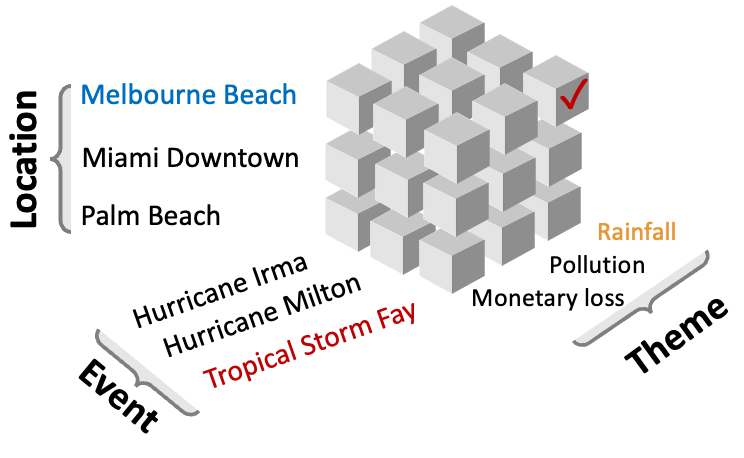}
  \caption{Access one cube cell in one hypercube. \textcolor{red!90}{\bf \checkmark} represents the touched cube cells.} 
  \label{fig:one_cube_one_cell}
\end{figure}

\subsection{Case 2: long query with multiple themes}
In cases where a query is very diverse, they may need to access multiple cube cells such that the query information can be covered as much as possible.
For example, the diverse query could be \emph{``What consequences were caused by Tropical Storm Fay, such as how much rainfall Melbourne Beach and Palm Bay in Florida received, and how much monetary losses were caused?''}.
Those cube cells needed could be 
<\textcolor{black}{Melbourne Beach}, \textcolor{black}{rainfall}, \textcolor{black}{Tropical Storm Fay} >,
<\textcolor{black}{Palm Beach}, \textcolor{black}{rainfall}, \textcolor{black}{Tropical Storm Fay} >,
<\textcolor{black}{Melbourne Beach}, \textcolor{black}{monetary losses}, \textcolor{black}{Tropical Storm Fay} >,
<\textcolor{black}{Palm Beach}, \textcolor{black}{monetary losses}, \textcolor{black}{Tropical Storm Fay}>. 
% <\textcolor{blue}{Melbourne Beach}, \textcolor{yellow!90}{rainfall}, \textcolor{red!90}{Tropical Storm Fay} >,
% <\textcolor{blue}{Palm Beach}, \textcolor{yellow!90}{rainfall}, \textcolor{red!90}{Tropical Storm Fay} >,
% <\textcolor{blue}{Melbourne Beach}, \textcolor{yellow!90}{monetary losses}, \textcolor{red!90}{Tropical Storm Fay} >,
% <\textcolor{blue}{Palm Beach}, \textcolor{yellow!90}{monetary losses}, \textcolor{red!90}{Tropical Storm Fay}>. 
%
\begin{figure}[H]
\centering  \includegraphics[width=0.5\columnwidth]{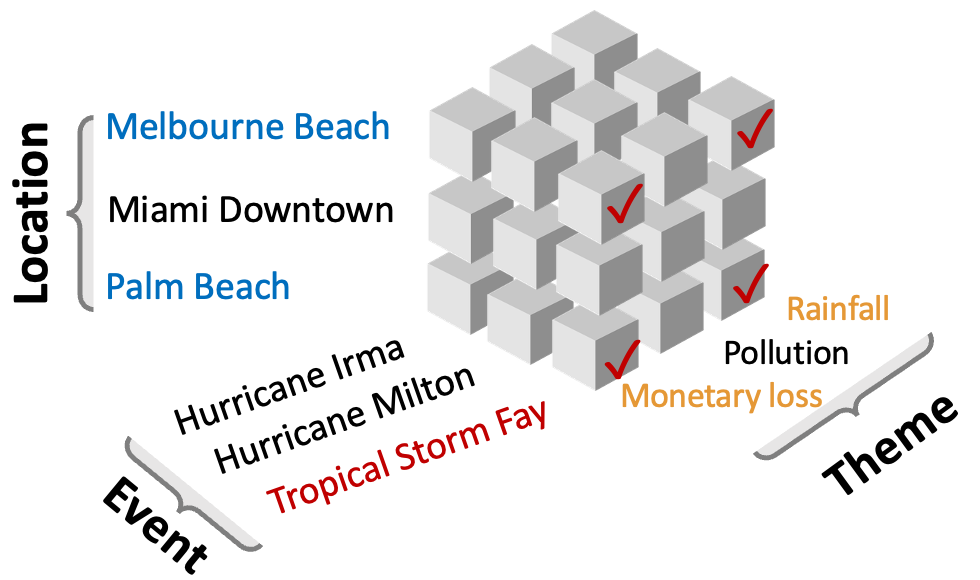}
  \caption{Access multiple cube cells in one hypercube. \textcolor{red!90}{\bf \checkmark} represents the touched cube cells.} 
  \label{fig:one_cube_multi_cells}
\end{figure}

\subsection{Case 3: diverse query with shifted themes} 
Given a more complicated query including shifted topics, e.g., \textit{``How did the monetary loss caused by Tropical Storm Fay impact the industrial layoff in Miami Downtown?''}, cube cells across two 
hypercubes are needed 
<\textcolor{black}{Miami Downtown}, \textcolor{black}{Industrial Layoff}, \textcolor{black}{Tropical Storm Fay} > and 
<\textcolor{black}{Miami Downtown}, \textcolor{black}{Monetary loss}, 
\textcolor{black}{Tropical Storm Fay} > to get more relevant documents.
These cross-theme queries are usually complex and need more dedicated hypercube design, which is considered for future exploration.
% <\textcolor{blue}{Miami Downtown}, \textcolor{yellow!90}{Industrial Layoff}, \textcolor{red!90}{Tropical Storm Fay} > and 
% <\textcolor{blue}{Miami Downtown}, \textcolor{yellow!90}{Monetary loss}, 
% \textcolor{red!90}{Tropical Storm Fay} > to get more relevant documents.
%
\begin{figure*}[ht]
\centering  \includegraphics[width=\columnwidth]{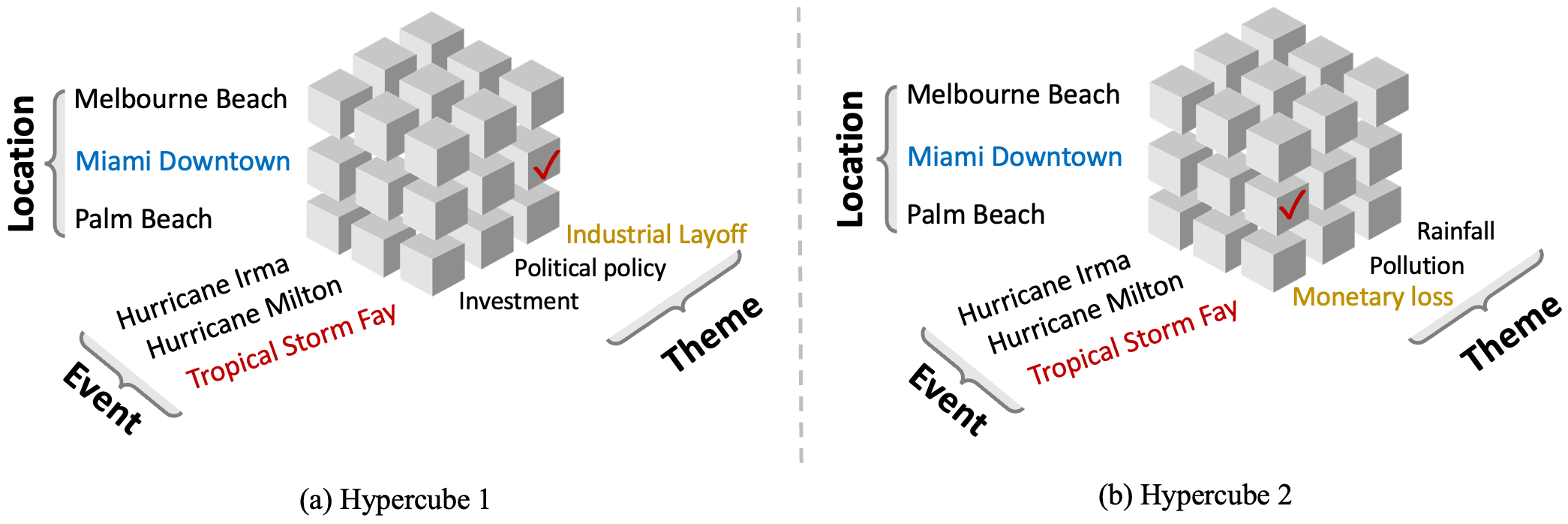}
  \caption{Access multiple cube cells in multiple hypercubes. \textcolor{red!90}{\bf \checkmark} represents the touched cube cells. } 
  \label{fig:two_cubes}
\end{figure*}

\end{document}